\documentclass{article} %
\usepackage{iclr2026_conference,times}
\iclrfinalcopy

\usepackage{amsmath,amsfonts,bm}

\def\eqref#1{equation~\ref{#1}}

\def\1{\bm{1}}

\DeclareMathAlphabet{\mathsfit}{\encodingdefault}{\sfdefault}{m}{sl}
\SetMathAlphabet{\mathsfit}{bold}{\encodingdefault}{\sfdefault}{bx}{n}

\usepackage{wrapfig}
\usepackage{hyperref}
\usepackage{url}
\usepackage{graphicx}
\usepackage{subcaption}
\usepackage{booktabs}
\usepackage{multirow}
\usepackage{booktabs, makecell, multirow}
\usepackage{caption}
\usepackage[table]{xcolor}

\usepackage{enumitem}
\setitemize{noitemsep,topsep=0pt,parsep=0pt,partopsep=0pt}

\usepackage{makecell}
\title{\raisebox{-0.5em}{\includegraphics[height=1.8em]{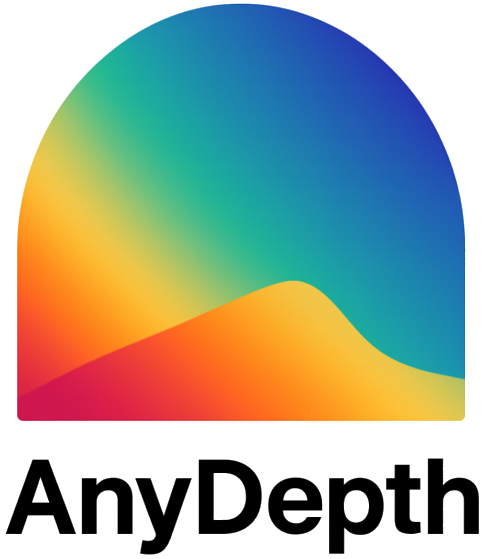}}~AnyDepth: Depth Estimation Made Easy}

\author{Zeyu Ren$^{1*}$~~
Zeyu Zhang$^{2*\dag}$~~
Wukai Li$^{2}$~~
Qingxiang Liu$^{3}$~~
Hao Tang$^{2\ddag}$\\[0.5em]
$^1$The University of Melbourne~~
$^2$Peking University~~
$^3$Shanghai University of Engineering Science\\[0.3em]
\footnotesize $^*$Equal contribution.
$^\dag$Project lead.
$^\ddag$Corresponding author: bjdxtanghao@gmail.com.
}

\begin{document}

\maketitle

\vspace{-0.8cm}
\begin{center}
    \textit{“Simplicity is prerequisite for reliability.” --- Edsger W. Dijkstra}
\end{center}

\begin{figure}[h]
\centering
\includegraphics[width=\textwidth]{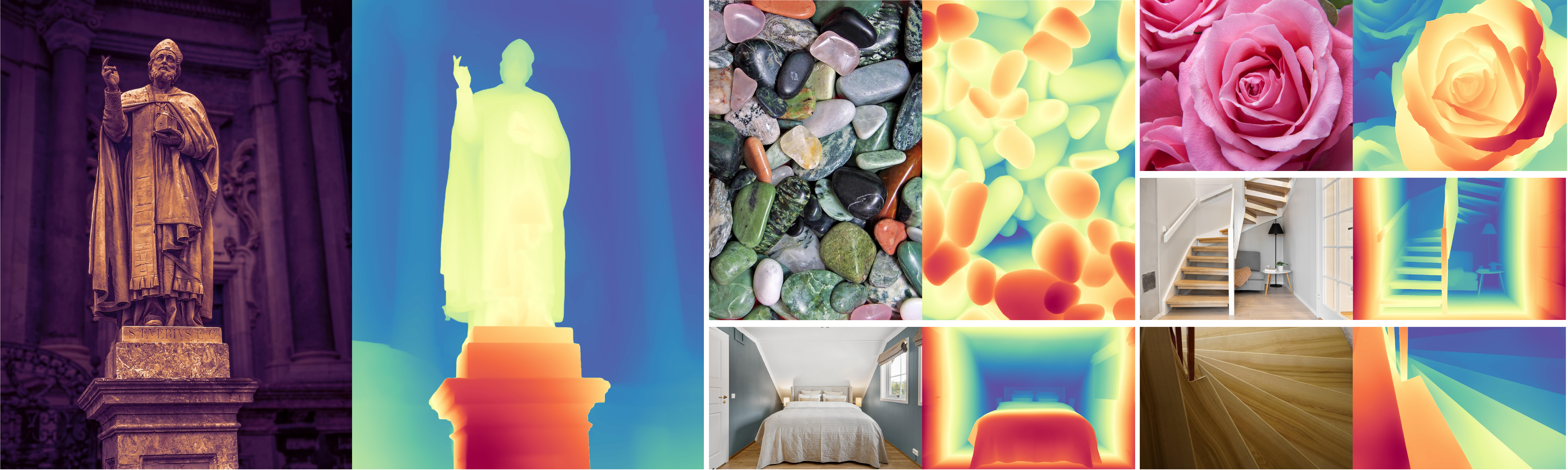}
\caption{We present \textbf{AnyDepth}, a simple and efficient training framework for zero-shot monocular depth estimation, which achieves impressive performance across a variety of indoor and outdoor scenes.}
\label{fig:teaser}
\end{figure}

\begin{abstract}

Monocular depth estimation aims to recover the depth information of 3D scenes from 2D images. Recent work has made significant progress, but its reliance on large-scale datasets and complex decoders has limited its efficiency and generalization ability. In this paper, we propose a lightweight and data-centric framework for zero-shot monocular depth estimation. We first adopt DINOv3 as the visual encoder to obtain high-quality dense features. Secondly, to address the inherent drawbacks of the complex structure of the DPT, we design the Simple Depth Transformer (SDT), a compact transformer-based decoder. Compared to the DPT, it uses a single-path feature fusion and upsampling process to reduce the computational overhead of cross-scale feature fusion, achieving higher accuracy while reducing the number of parameters by approximately 85\%–89\%. Furthermore, we propose a quality-based filtering strategy to filter out harmful samples, thereby reducing dataset size while improving overall training quality. Extensive experiments on five benchmarks demonstrate that our framework surpasses the DPT in accuracy. This work highlights the importance of balancing model design and data quality for achieving efficient and generalizable zero-shot depth estimation.
Code: \url{https://github.com/AIGeeksGroup/AnyDepth}.
Website: \url{https://aigeeksgroup.github.io/AnyDepth}.
\end{abstract}

\section{Introduction}

\begin{figure}[htbp]
    \centering
    \begin{subfigure}[t]{0.49\textwidth}
        \centering
        \includegraphics[width=\linewidth]{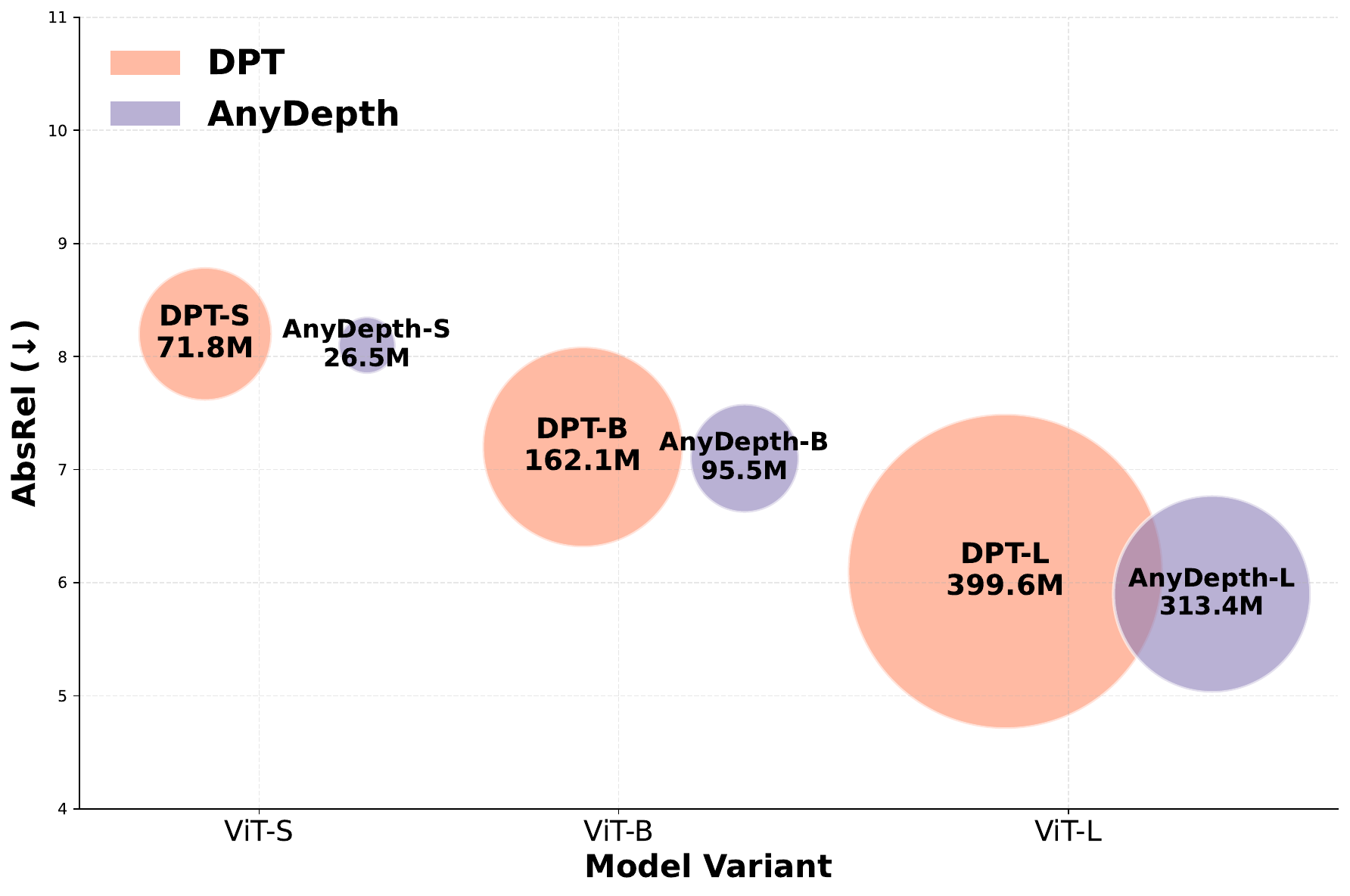}
        \caption{Model Comparison}
        \label{fig:model}
    \end{subfigure}
    \hfill
    \begin{subfigure}[t]{0.49\textwidth}
        \centering
        \includegraphics[width=\linewidth]{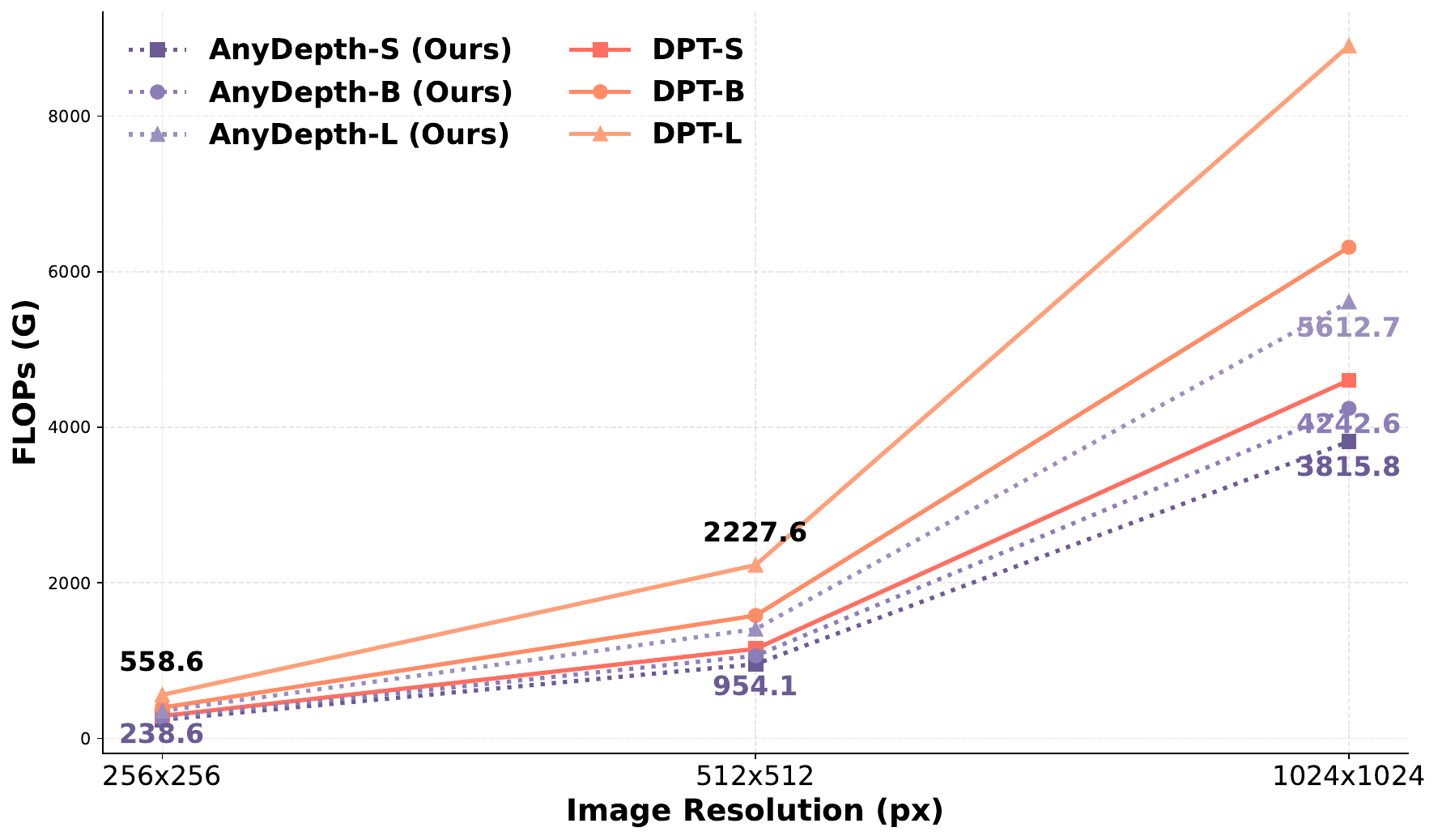}
        \caption{FLOPs Comparison}
        \label{fig:flops}
    \end{subfigure}
    \caption{Comparison of the number of parameters (left) and computational complexity (right) of \textbf{AnyDepth} and DPT for different model sizes and input resolutions. Our method significantly reduces the number of model parameters and computational cost while maintaining competitive accuracy.}
    \label{fig:model_flops}
    \vspace{-0.4cm}
\end{figure}

Monocular depth estimation is gaining increasing attention due to its wide range of downstream applications. Depth maps are not only used to measure scene distances \citep{Zoedepth,adabins,godard2017unsupervised}, but can also be embedded as conditional information within models in the 3D reconstruction~\citep{wang2025volsplat,wang2025drivegen3d,wang2025zpressor}, generation~\citep{controlnet,LDM,dreamfusion,nerf,Dngaussian,freenerf}, and embodied AI~\citep{wu2025stereoadapter,huang2025mobilevla,liu2025evovla,liu2025nav,huang20253d,song2025maniplvm,ye2025vla,huang20253dcoca,huang2025dc}, providing complementary information to improve granularity and geometric consistency.The MiDaS series \citep{MiDaS,midas3.1}, through extensive and systematic experiments, compared the transfer performance of various pretrained vision transformers (such as ViT \citep{vit}, Swin \citep{swin}, DINO \citep{dinov2}, and BeiT \citep{beit}) on monocular depth estimation tasks. DPT \citep{DPT} has demonstrated impressive performance in various dense prediction tasks and is currently used as the decoder in mainstream models. DPT aims to achieve finer-grained predictions by fusing features at different scales. The Depth Anything series \citep{da1,da2} represents a typical data-driven approach, aiming to improve  understanding and generalization capabilities of model for complex scenarios by leveraging massive datasets. These methods have significantly improved performance in zero-shot scenarios, demonstrating the potential of data scalability in the field of depth estimation. 

\begin{wrapfigure}{r}{0.5\textwidth} %
  \centering
  \includegraphics[width=0.48\textwidth]{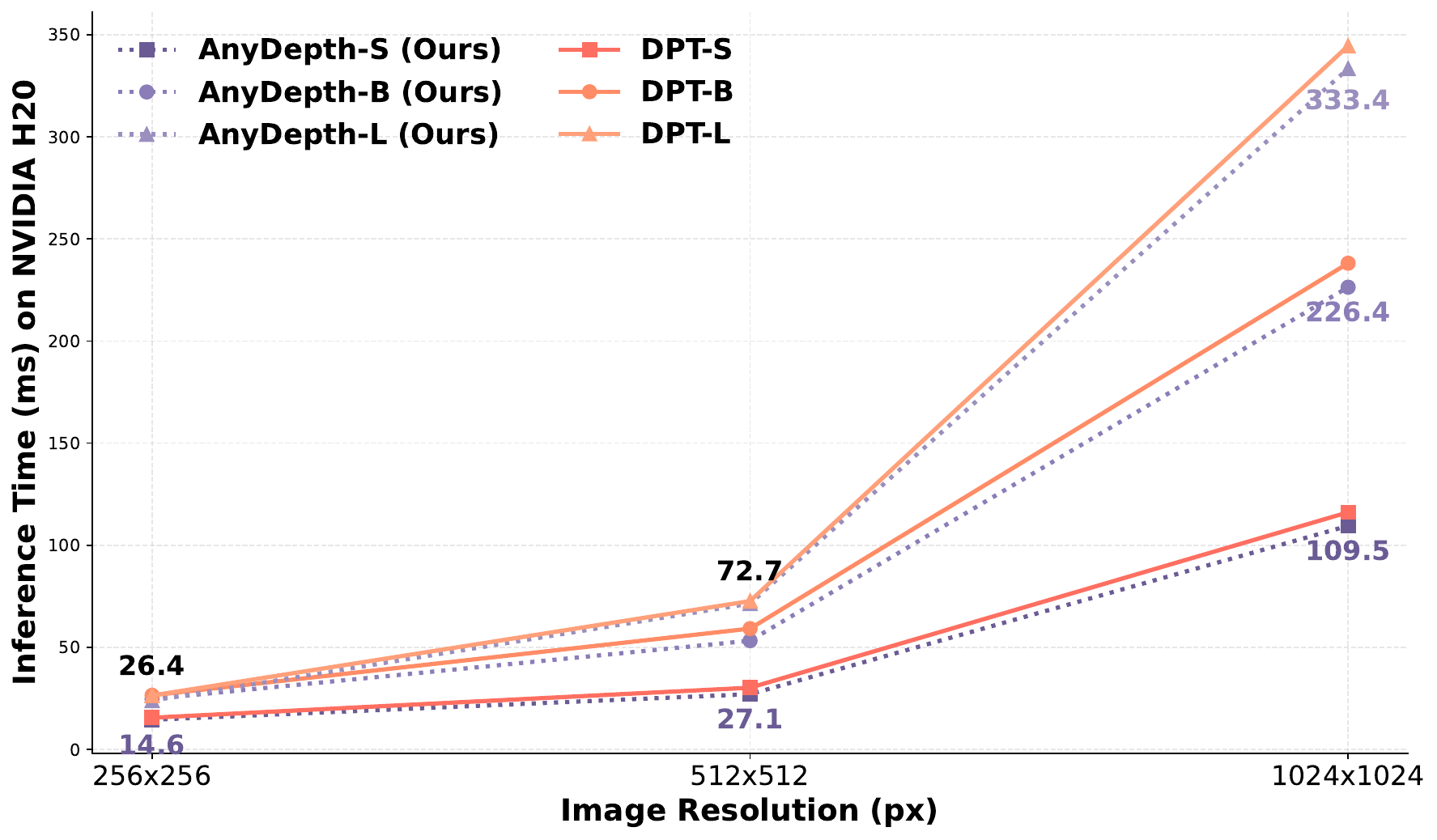}
  \caption{Comparison of inference time between AnyDepth and DPT at different input resolutions. Our method consistently achieves lower latency, especially at higher resolutions.}
  \label{fig:inference_time}
\end{wrapfigure}

However, We rethink the monocular depth estimation pipeline from both architectural and data-centric perspectives. From the architectural perspective, we observe that each Transformer layer in DPT requires a dedicated Reassemble module to map features to different scales, followed by multiple alignment operations. This design introduces unnecessary complexity, large parameter counts, and slow inference speed. DPT uses fixed bilinear interpolation for upsampling, which lacks adaptability to local geometric structures and often leads to blurred edges and loss of fine spatial details. From the data perspective, purely data-driven approaches such as the Depth Anything series rely heavily on massive datasets. However, large-scale data collection is costly and inevitably introduces noisy samples that degrade training quality. Simply scaling model size and data quantity therefore provides limited gains and poor reproducibility.

Based on these findings and limitations, we aim to design a lightweight and efficient training framework that maintains competitive performance while being widely adopted by the research community (Fig. \ref{fig:model_flops}).

Specifically, our contributions are reflected in three aspects:

\begin{itemize}
    
    \item We design a novel decoder that aligns and fuses features before restoring resolution through a one-shot reconstruction and upsampling. This architecture avoids multi-branch cross-scale alignment and repeated reconstruction, better preserving high-frequency details and geometric consistency.
    
    \item We analyze sample quality issues in deep learning datasets and proposed two metrics to quickly measure sample quality, which we then used to filter out low-quality samples. This reduced dataset size while improving overall data quality, demonstrating that our framework can achieve better performance with fewer resources.
    
    \item On multiple benchmarks, our framework achieves comparable accuracy and generalization to DPT with significantly fewer parameters and lower training overhead, demonstrating a superior efficiency-accuracy trade-off and academic reproducibility.
\end{itemize}

\section{Related Work}
\paragraph{Zero-Shot Monocular Depth Estimation.}
To enable widespread use of depth images in real-world scenarios without relying on specific environments, zero-shot depth estimation has become a key research direction in recent years~\citep{single_depth_perception,UniDepth,Oasis,Learning_to_recover3d}. Due to the lack of strict geometric constraints on MDE, many zero-shot models learn to predict affine-invariant depth, i.e., recovering relative structure while maintaining scale and translation invariance~\citep{MiDaS,da1,da2}. For example, DiverseDepth~\citep{diversedepth} uses web images as training data to improve zero-shot generalization performance. MiDaS~\citep{MiDaS} proposed scale-shift-invariant losses to solve the ambiguity problem of different deep numerical representation methods of different datasets, so that the model can be trained on a large scale. In order to eliminate the inherent problems of the CNN backbone, the performance of Zero-Shot Monocular Depth Estimation was further improved by using the vision transformer architecture, such as DPT~\citep{DPT}, Omnidata~\citep{Omnidata}, Depthformer~\citep{Depthformer} and Zoepdeth~\citep{Zoedepth}. Marigold~\citep{Marigold} directly utilizes the standard diffusion model paradigm and stable diffusion pre-trained weights for fine-tuning to produce high-quality results. Depth Anything series~\citep{da1,da2} used 62 million unlabeled images for larger-scale training. Geowizard~\citep{Geowizard} uses the high consistency between dense prediction tasks to jointly predict depth and normals. Lotus~\citep{Lotus} analyzes the diffusion process to achieve single-step diffusion and speed up the inference process. Genpercept~\citep{Genpercept} uses experiments to prove that the diffusion model requires specific details to be optimized in dense prediction tasks.

\paragraph{Decoder for Dense Prediction.}
Currently, many methods for dense prediction tasks employ multi-scale feature fusion strategies to compensate for the lack of information from single-layer features~\citep{FPN,PAN,efficientdet,encoder-decoder,Nas-fpn,xu2021monocular,eigen2015predicting}. FPN~\citep{FPN} proposes a top-down architecture where high-level semantic representations are successively merged with low-level features to enhance multi-scale features.~\citep{From_big_to_small} designed a multi-scale local plane guidance layer to more effectively guide the fusion of features at each layer to achieve performance improvement. Swin-Depth~\citep{swin-depth} designs a lightweight multi-scale attention mechanism module to enhance the ability to learn global information at multiple scales. PVT~\citep{PVT} and Uformer~\citep{Uformer} use a multi-scale pyramid decoder structure to capture long-range visual dependencies.DPT~\citep{DPT} utilizes the ViT~\citep{vit} backbone network to generate high-resolution features, thereby achieving finer-grained representation and improving prediction accuracy. However, multi-branch reassembly incurs significant overhead, especially in the case of high-resolution input.
\section{The Proposed Method}

\subsection{Overview}
\begin{figure}
\centering
\includegraphics[width=\textwidth]{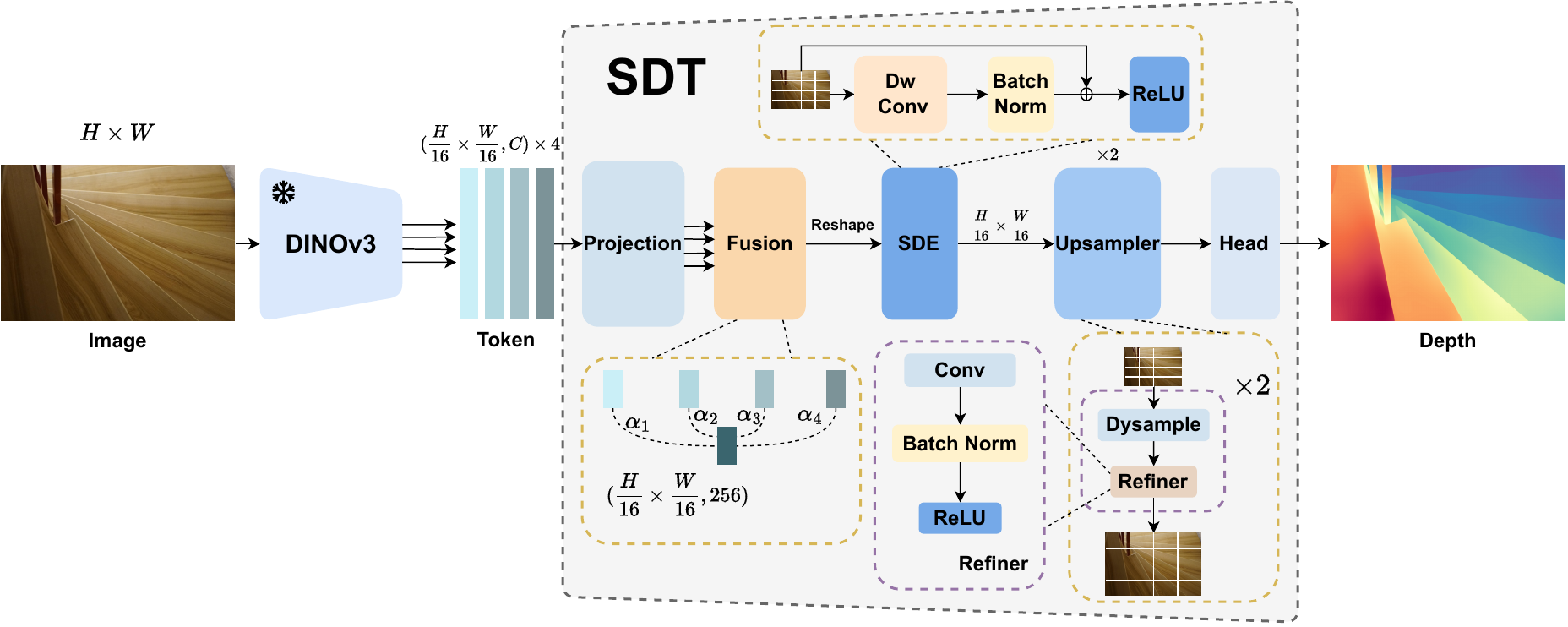}
\caption{\textbf{\textbf{AnyDepth} architecture overview.} The input image is encoded into tokens by a frozen DINOv3 backbone network, then decoded by our lightweight SDT decoder. Tokens undergo only a single projection and weighted fusion. The Spatial Detail Enhancer (SDE) module ensures finer-grained predictions. The feature map is upsampled by an efficient and learnable upsampler dysample, and the depth is finally output by the head.}
\label{fig:overview}
    \vspace{-0.4cm}
\end{figure}

The proposed AnyDepth uses a pre-trained DINOv3 \citep{dinov3} encoder and SDT decoder; as shown in Fig.\ref{fig:overview}, given an input image $I$, we extract multi-scale representations from four intermediate Transformer layers ${T^1, T^2, T^3, T^4}$ and input them into the SDT head for depth reconstruction, thereby capturing different levels of detail and semantic information. These tokens are linearly projected onto a common dimension and fused to capture complementary semantic levels. The fused representations are then reshaped into feature maps and refined by a Spatial Detail Enhancer (SDE). Finally, a dense depth map is generated through two learnable Upsampler and head prediction.Our method differs from the Depth Anything series~\citep{da1,da2} and DPT~\citep{DPT} in that we fuse tokens using only a single linear projection, followed by upsampling in a single path, without multi-branch cross-scale alignment, significantly reducing the number of parameters and computational overhead.

\subsection{Simple Depth Transformer (SDT)}
Our decoder adopts a simple single-path fusion and reconstruction strategy, aiming to take advantage of the high-resolution feature of DINOv3 and further unleash its performance at high resolution. We first project the tokens extracted from the encoder into a 256-dimensional space using a linear layer followed by a GELU non-linearity \citep{gelus}, which preserves sufficient informative content while substantially reducing the computational overhead in the subsequent decoding stages. For the class token, we keep the same processing as DPT \citep{DPT}, concatenate it with the spatial token, and then fuse it through the learnable projection.

\paragraph{Fusion.} To fuse tokens from multiple layers of representation, we then employ a learnable weighted fusion strategy (Eq.~\ref{eq:fusion}). 

Specifically, we assign a learnable scalar weight to each layer of tokens and normalize them using a softmax function to form a uniform probability distribution, preventing initial instability in training. This strategy enables the model to adaptively balance low-level structural details with high-level semantic information.
\begin{equation}
    T = \sum_{i \in \mathcal{L}} \alpha_i \, \mathrm{Proj}_i(T_i), 
    \quad T_i \in \mathbb{R}^{N_p \times D},
    \label{eq:fusion}
\end{equation}
Where $T_i$ denotes the token in layer $i$ after projection, and contains $N_p$ tokens of dimension $D$.

\paragraph{Spatial Detail Enhancer.}
After the fusion block, we reshape the sequence token output into a spatial feature map. Because the reorganized feature map lacks local continuity and, after multi-level fusion, easily obscures shallow texture details, which are crucial for dense prediction tasks such as depth estimation, we designed the Spatial Detail Enhancer.The SDE can be expressed by Eq.~\ref{eq:sde},
\begin{equation}
    F' = ReLU(F + BN(DWConv_{3\times3}(F))),  \
    F \in \mathbb{R}^{\frac{H}{16}\times\frac{W}{16} \times 256}.
    \label{eq:sde}
\end{equation}
We implement this operation first using a $3\times3$ Depthwise convolution for local spatial modeling, followed by batch normalization. We then add the normalized response to the input feature $F$ via a residual connection, and finally pass it through an activation layer.

\paragraph{Upsampler.}
In the upsampling stage, we abandon the commonly used bilinear interpolation, which easily blurs high-frequency details, and instead adopt a learnable dynamic sampler (Eq.~\ref{eq:Upsample}). Specifically, we use DySample~\citep{dysampler} as the upsampler, which adaptively constructs an offset sampling grid based on the learned low-resolution features to adjust the sampling position, and then uses differentiable grid sampling to resample to high-resolution features. We first define three operators: the DySample block $\mathcal{B}(\cdot)$, the DySample stage $\mathcal{S}(\cdot)$, and the refinement block $\mathcal{R}(\cdot)$:
\begin{equation}
\mathcal{B}(X) 
= \mathrm{ReLU}\!\Big(\mathrm{BN}\big(\mathrm{Conv}_{3\times3}(\mathrm{DySample}_{\times 2}(X))\big)\Big),
\label{eq:DySampleBlock}
\end{equation}
\begin{equation}
\mathcal{S}(X) = \mathcal{B}\!\big(\mathcal{B}(X)\big),
\label{eq:DySampleStage}
\end{equation}
\begin{equation}
\mathcal{R}(X) 
= \mathrm{ReLU}\!\Big(\mathrm{BN}\big(\mathrm{Conv}_{3\times3}(X)\big)\Big).
\label{eq:Refine}
\end{equation}

Based on these definitions (Eq.~\ref{eq:DySampleBlock}, \ref{eq:DySampleStage}, \ref{eq:Refine}), the complete upsampling process can be expressed as:
\begin{equation}
\mathcal{U}(X) 
= \mathcal{R}\!\Big(\mathcal{S}\big(\mathcal{R}(\mathcal{S}(X))\big)\Big),
\label{eq:Upsample}
\end{equation}
In this way, the compact feature map of size $H/16 \times W/16$ can be progressively upsampled back to the original resolution $H \times W$. We want to emphasize that we do not jump to $H\times W$ all at once, but rather decompose the upsampling into two $\times4$ upsamplers, using four dysamples of scale 2. Single-stage $\times16$ upsampling forces the sampler to infer large offsets from very low-resolution features, which amplifies errors and destabilizes gradients. Our progressive design keeps the offsets small, inserting local refinement after each resampling, resulting in a model with better detail recovery capabilities.

\subsection{SDT vs. DPT}
A key difference between SDT and DPT \citep{DPT} is the order of feature reassembly. DPT employs a reassemble-fusion strategy. Specifically, DPT first applies the reassemble module to the tokens extracted by each Transformer layer, mapping the tokens to feature maps of different scales. These feature maps are then fused in a cascade across scales, which inevitably introduces multiple branches and repeated cross-scale alignment overhead. In contrast, SDT employs a fusion-reassemble strategy, directly projecting and fusing groups of tokens. Only after this stage do we perform spatial reassembly and upsampling along a single path. This fusion-reassemble strategy avoids the high cost of per-layer token reassembly and feature map cross-scale alignment, making it more efficient and stable, especially when processing high-resolution inputs.

\section{Experiments}

\subsection{Datasets and Metrics}
\paragraph{Training Datasets.}
We use five synthetic datasets covering various indoor and outdoor scenes for training.
(1) \emph{Hypersim}~\citep{hypersim} after filtering incomplete samples, we have approximately $39$K.
(2) \emph{Virtual KITTI}~\citep{vkitti} we selected four scenes, totaling approximately $20$K.
(3) \emph{BlendedMVS}~\citep{blendedmvs} 
(4) \emph{IRS}~\citep{irs} 
(5) \emph{TartanAir}~\citep{tartanair}  As shown in Table \ref{tab:dataset_stats}, we only use 369K datasets for training.
The far plane is set to $100\,\mathrm{m}$. To improve the robustness and generalization of the model, we used data augmentation of flipping and rotation.

\paragraph{Evaluation Datasets and Metrics.}
For Zero-shot monocular depth estimation, we evaluate SDT using five datasets containing various scenes: NYUv2 \citep{nyuv2}, KITTI \citep{kitti}, ETH3D \citep{eth3d}, ScanNet \citep{scannet}, and DIODE \citep{diode}. We use the absolute mean relative error(AbsRel), i.e., $\frac{1}{M}\sum_{i=1}^{M}\frac{|\hat d_i-d_i|}{d_i}$, where $M$ is the total number of valid pixels, $d_i$ denotes the ground truth, and $\hat d_i$ is the predicted depth. We report accuracy thresholds $\delta_\tau$, which denote the fraction of pixels where the prediction and ground truth differ by less than a multiplicative factor $\tau=1.25$.

\subsection{Implementation Details}

Our setup differs slightly from Depth Anything V2 \citep{da2}. To better utilize the high-resolution features of DINOv3~\citep{dinov3}, we increase the input image resolution to \(768\times768\). The encoder is kept frozen throughout training, and we use features from four intermediate layers as decoder inputs: \([2, 5, 8, 11]\) for DINOv3 S/16 and DINOv3 B/16, and \([4, 11, 17, 23]\) for DINOv3 L/16. We perform simple regression to predict disparity $d' = 1 / d$, where \(d'\) denotes disparity and \(d\) denotes depth. Both the input image and the groundtruth are normalized to \([0,1]\). We follow the settings of Depth Anything v2 \citep{da2} and use a scale- and shift-invariant loss $\mathcal{L}_{\mathrm{ssi}}$ and a gradient matching loss $\mathcal{L}_{\mathrm{gm}}$, and the weight ratio of $\mathcal{L}_{\mathrm{ssi}}$ and $\mathcal{L}_{\mathrm{gm}}$ is set to $1:2$.
To stabilize optimization, we follow an optimization strategy similar to DINOv3 \citep{dinov3}. We use AdamW with a base learning rate of \(1\times10^{-3}\), a PolyLR scheduler with power \(0.9\), and a linear warm-up for the first two epochs. We train for a total of five epochs.

\subsection{Main Results}

\subsubsection{Results of Data Centric Learning}

\begin{figure}[t]
  \centering
  \begin{subfigure}{0.32\textwidth}
    \centering
    \includegraphics[width=\linewidth]{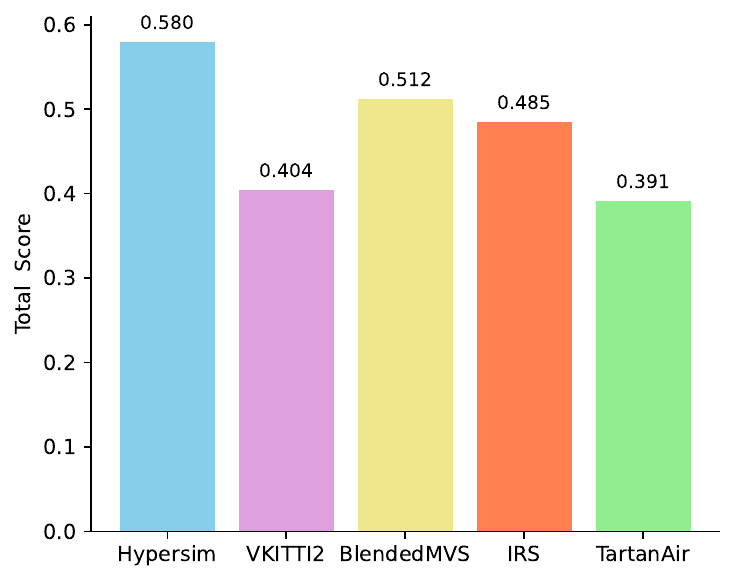}
    \caption{Total Score}
  \end{subfigure}
  \hfill
  \begin{subfigure}{0.32\textwidth}
    \centering
    \includegraphics[width=\linewidth]{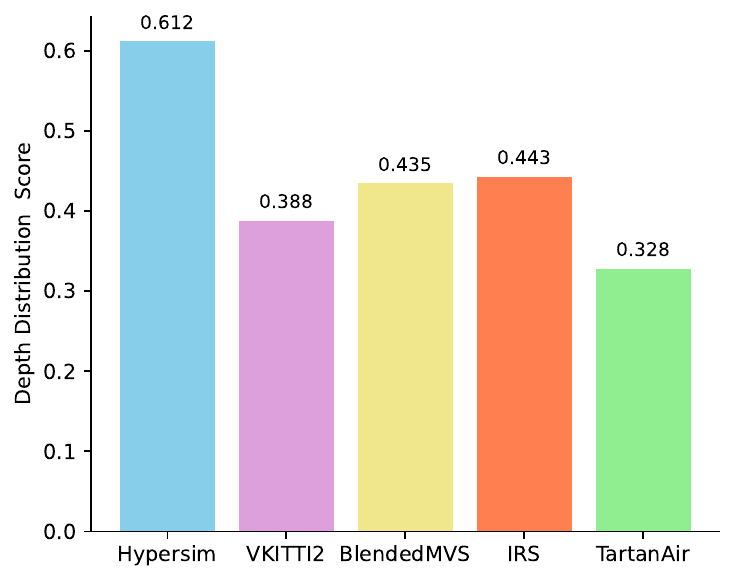}
    \caption{Depth Distribution Score}
  \end{subfigure}
  \hfill
  \begin{subfigure}{0.32\textwidth}
    \centering
    \includegraphics[width=\linewidth]{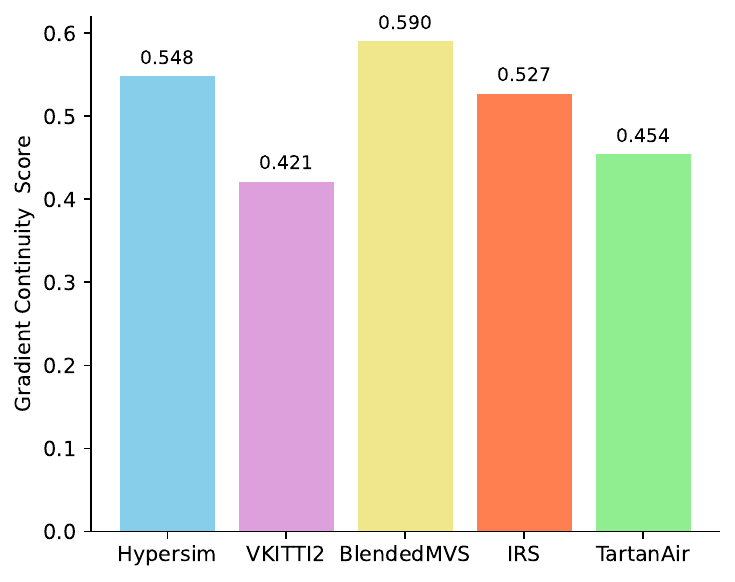}
    \caption{Gradient Continuity Score}
  \end{subfigure}
  \caption{Dataset quality across the Total Score, Depth Distribution Score, and Gradient Continuity Score (higher is better).}
  \label{fig:metrics_barplots}
\end{figure}

\begin{wraptable}{r}{0.45\textwidth}
\vspace{-8pt}
\centering
\caption{Dataset statistics of good and bad samples.}
\label{tab:dataset_stats}
\small
\setlength{\tabcolsep}{6pt}
\renewcommand{\arraystretch}{1.15}
\begin{tabular}{lrrr}
\toprule
\textbf{Dataset} & \textbf{Total} & \textbf{Good} & \textbf{Bad} \\
\midrule
Hypersim   & 39,648  & 26,912  & 12,736  \\
VKITTI2    & 19,559  & 12,643  & 6,916   \\
BlendedMVS & 115,142 & 74,838  & 40,304  \\
IRS        & 103,316 & 68,211  & 35,105  \\
TartanAir  & 306,637 & 186,693 & 119,944 \\
\midrule
\textbf{Summary} & 584,302 & 369,297 & 215,005 \\
\bottomrule
\end{tabular}
\vspace{-6pt}
\end{wraptable}

We applied the metrics proposed in Section \ref{sec:Data centric learning} to all training datasets, with the results shown in Fig. \ref{fig:metrics_barplots}. We observe that Hypersim performed well in both the Depth Distribution Score and Gradient Continuity Score, achieving the highest overall score. This indicates a relatively balanced depth distribution, smooth gradients, and a low concentration of noisy samples. In contrast, datasets containing outdoor samples, such as VKITTI2, BlendedMVS, and TartanAir, had significantly lower Depth Distribution Scores, indicating a more severe depth distribution. This is likely a common problem across all outdoor datasets. The low Gradient Continuity Score for VKITTI2 may be due to the presence of numerous fine-grained structures (\emph{e.g.}, leaves) in the samples, resulting in abundant edges and severe gradient abruptness, which is considered noisy.

Following the methods described in Section \ref{sec:Data centric learning}, we filtered the entire dataset. Specifically, we first filtered out samples whose valid depth values accounted for less than 20\% of the total pixels. We then sorted the remaining samples based on the Depth Distribution Score and Gradient Continuity Score, filtering out the 20\% with the lowest scores for each metric. The number of filtered samples for each dataset is shown in Table~\ref{tab:dataset_stats}. For visualizations of low-quality samples, please see the \ref{Visualization of low-quality samples}. The merged dataset contains 584K samples, of which approximately 369K are used for training and 215K are filtered out.

\subsubsection{QUANTITATIVE COMPARISONS}

\begin{table}[t]
\centering
\caption{Quantitative comparison of zero-shot affine-invariant depth estimation. Lower AbsRel values are better; higher $\delta_1$ values are better. DINOv3 \citep{dinov3} uses the ViT-7B encoder, and Depth Anything v2 (DAv2) \citep{da2} is trained on 62.6M datasets. For fair comparison, the baseline (DPT) uses a frozen DINOv3 encoder and DPT head, while our method replaces the DPT head with the proposed SDT. The bold numbers in the table refer to the best results between DPT and AnyDepth.}
\label{tab:quant_affine}
\small
\setlength{\tabcolsep}{0.7pt}
\renewcommand{\arraystretch}{0.9}
\begin{tabular}{l l l c *{5}{cc}}
\toprule
\multirow{2}{*}{\bf Method} & \multirow{2}{*}{\bf \makecell{Training \\ Data$\downarrow$}} & \multirow{2}{*}{\bf Encoder} & \multirow{2}{*}{\bf \makecell{\#Params \\ (M)$\downarrow$}} 
& \multicolumn{2}{c}{\bf NYUv2 }
& \multicolumn{2}{c}{\bf KITTI}
& \multicolumn{2}{c}{\bf ETH3D }
& \multicolumn{2}{c}{\bf ScanNet }
& \multicolumn{2}{c}{\bf DIODE } \\
\cmidrule(lr){5-6}\cmidrule(lr){7-8}\cmidrule(lr){9-10}\cmidrule(lr){11-12}\cmidrule(lr){13-14}
& & & & AbsRel$\downarrow$ & $\delta_1\uparrow$
  & AbsRel$\downarrow$ & $\delta_1\uparrow$
  & AbsRel$\downarrow$ & $\delta_1\uparrow$
  & AbsRel$\downarrow$ & $\delta_1\uparrow$
  & AbsRel$\downarrow$ & $\delta_1\uparrow$ \\
\midrule
DINOv3 & 595K & ViT-7B & 91.19 & 4.3 & 98.0 & 7.3 & 96.7 & 5.4 & 97.5 & 4.4 & 98.1 & 25.6 & 82.2 \\
\midrule
\multirow{3}{*}{DAv2} 
& \multirow{3}{*}{62.6M} & ViT\text{-}S & 71.8  & 5.3 & 97.3  & 7.8 & 93.6  & 14.2 & 85.1  & -- & --  & 7.3 & 94.2 \\
&                       & ViT\text{-}B & 162.1 & 4.9 & 97.6  & 7.8 & 93.9  & 13.7 & 85.8  & -- & --  & 6.8 & 95.0 \\
&                       & ViT\text{-}L & 399.6 & 4.5 & 97.9  & 7.4 & 94.6  & 13.1 & 86.5  & -- & --  & 6.6 & 95.2 \\

\midrule
\multirow{3}{*}{DPT} 
& \multirow{3}{*}{584K} & ViT\text{-}S & 71.8 & 8.4 & 93.3 & 10.8 & 89.1 & 12.7 & 92.0 & 8.3 & 93.5 & 26.0 & 71.4 \\
&                     & ViT\text{-}B & 162.1 & 7.5 & 95.1 & 10.8 & 88.9 & 10.0 & 92.9 & 7.1 & 95.3 & 24.5 & 73.4 \\
&                     & ViT\text{-}L & 399.6 & 6.1 & \textbf{96.8} & 8.9 & 92.5 & 13.0 & 94.9 & 6.0 & 97.0 & 23.4 & \textbf{73.9} \\
\midrule
\multirow{3}{*}{\bf AnyDepth} 
& \multirow{3}{*}{369K} & ViT\text{-}S & 26.5  & 8.2 & 93.2 & 10.2 & 88.3 & 8.4  & 93.5 & 8.0 & 93.6 & 24.7 & 71.4 \\
&                            & ViT\text{-}B & 95.5  & 7.2 & 95.0 & 9.7  & 90.1 & \textbf{8.0}  & 94.5 & 6.8 & 95.6  & 23.6 & 72.7  \\
&                            & ViT\text{-}L & 313.4 & \textbf{6.0}  &  \textbf{96.8} & \textbf{8.6}  & \textbf{92.6} & 9.6 & \textbf{95.4}  & \textbf{5.4} & \textbf{97.4}  & \textbf{22.6}  &  73.6 \\
\bottomrule
\end{tabular}
\end{table}

\begin{table}[t]
\centering
\caption{Comparison of zero-shot affine-invariant depth estimation with different encoders and decoders. Green cells indicate the best results within each method.}
\label{tab:quant_affine_decoder}
\small
\setlength{\tabcolsep}{1.2pt}
\renewcommand{\arraystretch}{0.92}
\begin{tabular}{l l l *{5}{cc}}
\toprule
\multirow{2}{*}{\bf Method} & \multirow{2}{*}{\bf Encoder} & \multirow{2}{*}{\bf Decoder}
& \multicolumn{2}{c}{\bf NYUv2 }
& \multicolumn{2}{c}{\bf KITTI}
& \multicolumn{2}{c}{\bf ETH3D }
& \multicolumn{2}{c}{\bf ScanNet }
& \multicolumn{2}{c}{\bf DIODE } \\
\cmidrule(lr){4-5}\cmidrule(lr){6-7}\cmidrule(lr){8-9}\cmidrule(lr){10-11}\cmidrule(lr){12-13}
& & 
& AbsRel$\downarrow$ & $\delta_1\uparrow$
& AbsRel$\downarrow$ & $\delta_1\uparrow$
& AbsRel$\downarrow$ & $\delta_1\uparrow$
& AbsRel$\downarrow$ & $\delta_1\uparrow$
& AbsRel$\downarrow$ & $\delta_1\uparrow$ \\
\midrule

\multirow{2}{*}{DAv2} 
& \multirow{2}{*}{ViT\text{-}B} & DPT 
& 5.8 & 96.2 
& \cellcolor[rgb]{.886,.937,.851}\textbf{10.4} & 89.1 
& 8.8 & 94.6 
& 6.2 & 95.3 
& \cellcolor[rgb]{.886,.937,.851}\textbf{23.4} & 73.8 \\
& & SDT 
& \cellcolor[rgb]{.886,.937,.851}\textbf{5.6} & \cellcolor[rgb]{.886,.937,.851}\textbf{96.4} 
& 10.7 & \cellcolor[rgb]{.886,.937,.851}\textbf{89.6} 
& \cellcolor[rgb]{.886,.937,.851}\textbf{7.5} & \cellcolor[rgb]{.886,.937,.851}\textbf{95.8} 
& \cellcolor[rgb]{.886,.937,.851}\textbf{6.1} & \cellcolor[rgb]{.886,.937,.851}\textbf{95.4} 
& 23.9 & \cellcolor[rgb]{.886,.937,.851}\textbf{73.9} \\
\midrule

\multirow{3}{*}{DAv3} 
& \multirow{3}{*}{ViT\text{-}L} & DPT
& \cellcolor[rgb]{.886,.937,.851}\textbf{4.9} & 96.9 
& \cellcolor[rgb]{.886,.937,.851}\textbf{8.8} & \cellcolor[rgb]{.886,.937,.851}\textbf{92.4} 
& 6.9 & 95.9 
& 5.0 & \cellcolor[rgb]{.886,.937,.851}\textbf{96.6} 
& 22.5 & 74.6 \\

& & Dual\text{-}DPT
& \cellcolor[rgb]{.886,.937,.851}\textbf{4.9} & 97.0 
& 8.9 & \cellcolor[rgb]{.886,.937,.851}\textbf{92.4} 
& 7.0 & 95.8 
& \cellcolor[rgb]{.886,.937,.851}\textbf{4.9} & \cellcolor[rgb]{.886,.937,.851}\textbf{96.6} 
& 22.3 & 74.6 \\

& & SDT
& \cellcolor[rgb]{.886,.937,.851}\textbf{4.9} & \cellcolor[rgb]{.886,.937,.851}\textbf{97.1} 
& 8.9 & \cellcolor[rgb]{.886,.937,.851}\textbf{92.4} 
& \cellcolor[rgb]{.886,.937,.851}\textbf{5.8} & \cellcolor[rgb]{.886,.937,.851}\textbf{96.6} 
& 5.0 & \cellcolor[rgb]{.886,.937,.851}\textbf{96.6} 
& \cellcolor[rgb]{.886,.937,.851}\textbf{21.9} & \cellcolor[rgb]{.886,.937,.851}\textbf{74.9} \\
\midrule

\multirow{2}{*}{VGGT} 
& \multirow{2}{*}{VGGT\text{-}1B} & DPT
& \cellcolor[rgb]{.886,.937,.851}\textbf{4.8} & 97.7 
& 15.6 & 77.9 
& 7.2 & 94.7 
& \cellcolor[rgb]{.886,.937,.851}\textbf{4.6} & 97.6 
& 30.7 & 76.2 \\

& & SDT
& \cellcolor[rgb]{.886,.937,.851}\textbf{4.8} & \cellcolor[rgb]{.886,.937,.851}\textbf{98.0} 
& \cellcolor[rgb]{.886,.937,.851}\textbf{15.5} & \cellcolor[rgb]{.886,.937,.851}\textbf{80.1} 
& \cellcolor[rgb]{.886,.937,.851}\textbf{7.0} & \cellcolor[rgb]{.886,.937,.851}\textbf{95.1} 
& \cellcolor[rgb]{.886,.937,.851}\textbf{4.6} & \cellcolor[rgb]{.886,.937,.851}\textbf{98.0} 
& \cellcolor[rgb]{.886,.937,.851}\textbf{30.6} & \cellcolor[rgb]{.886,.937,.851}\textbf{76.8} \\
\bottomrule
\end{tabular}
\end{table}

Table~\ref{tab:quant_affine} reports quantitative comparison results for zero-shot affine-invariant depth estimation. 
Since the baselines in the Depth Anything series all use a DPT head, we primarily compare our proposed SDT decoder with the DPT under the same backbone settings. 

\begin{figure}[t]
  \centering
  \includegraphics[width=\textwidth]{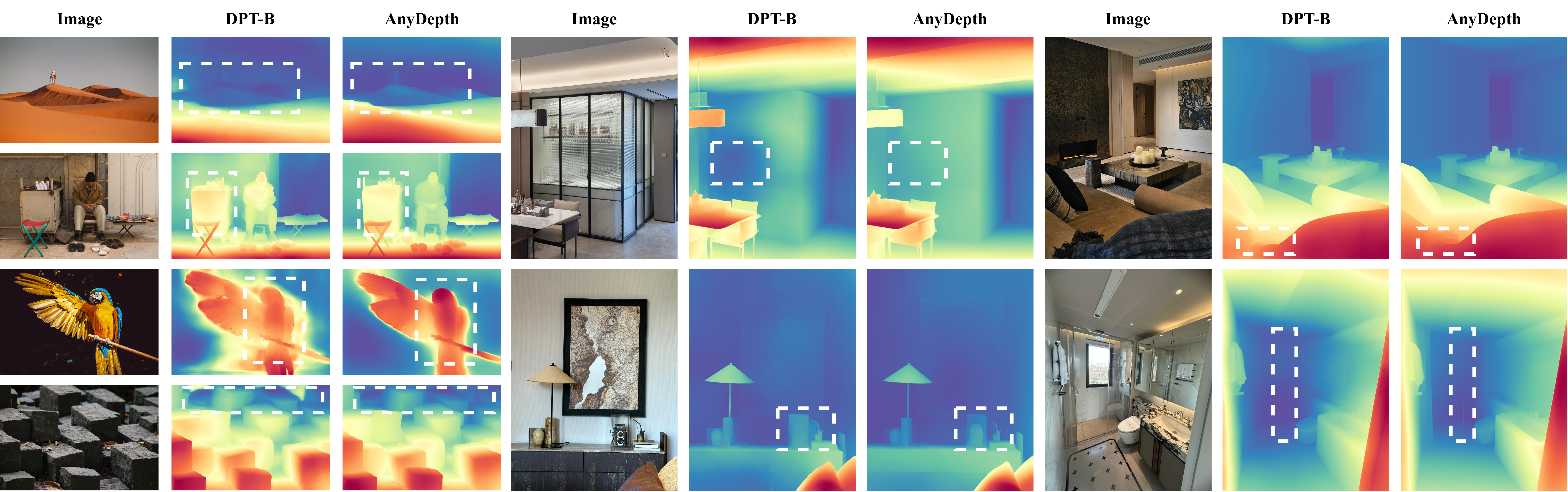}
  \caption{Qualitative results of zero-shot monocular depth estimation using \textbf{AnyDepth} of ViT-B and comparison with DPT-B.}
  \label{fig:gradients}
\end{figure}

\begin{table}[t]
\centering
\caption{Multi-resolution efficiency comparison of SDT and DPT heads under a ViT-L encoder. Latency is averaged over 1000 runs on an NVIDIA H100 GPU. Lower is better.}
\label{tab:efficiency_multi_res}
\begin{tabular}{l l r r}
\toprule
\textbf{Resolution} &
\textbf{Decoder} &
\textbf{FLOPs (G)$\downarrow$} &
\textbf{Latency (ms)$\downarrow$} \\
\midrule
\multirow{2}{*}{256$\times$256}
 & DPT & 444.14 & 6.66 $\pm$ 0.22 \\
 & \cellcolor[rgb]{.886,.937,.851}\textbf{SDT (Ours)}
 & \cellcolor[rgb]{.886,.937,.851}\textbf{234.17}
 & \cellcolor[rgb]{.886,.937,.851}\textbf{6.10 $\pm$ 0.33} \\
\midrule
\multirow{2}{*}{512$\times$512}
 & DPT & 1776.56 & 24.65 $\pm$ 0.22 \\
 & \cellcolor[rgb]{.886,.937,.851}\textbf{SDT (Ours)}
 & \cellcolor[rgb]{.886,.937,.851}\textbf{936.70}
 & \cellcolor[rgb]{.886,.937,.851}\textbf{23.17 $\pm$ 0.54} \\
\midrule
\multirow{2}{*}{1024$\times$1024}
 & DPT & 7106.22 & 99.79 $\pm$ 0.79 \\
 & \cellcolor[rgb]{.886,.937,.851}\textbf{SDT (Ours)}
 & \cellcolor[rgb]{.886,.937,.851}\textbf{3746.79}
 & \cellcolor[rgb]{.886,.937,.851}\textbf{93.09 $\pm$ 0.51} \\
\bottomrule
\end{tabular}
\end{table}

While our approach does not yet surpass the state-of-the-art results reported by fully data-driven methods (\emph{e.g.}, the Depth Anything series \citep{da1,da2} and DINOv3-7B \citep{dinov3}, which require hundreds of millions of parameters or massive datasets), 
\begin{wraptable}{r}{0.45\textwidth}
\vspace{-8pt}
\centering
\caption{Decoder parameter comparison across different ViT backbones. Lower is better.}
\label{tab:decoder_params_vit}
\setlength{\tabcolsep}{6pt}
\renewcommand{\arraystretch}{1}
\begin{tabular}{l l r}
\toprule
\textbf{Decoder} &
\textbf{ViT Backbone} &
\textbf{Params (M)$\downarrow$} \\
\midrule
\multirow{3}{*}{DPT}
 & ViT-S & 50.83 \\
 & ViT-B & 76.05 \\
 & ViT-L & 99.58 \\
\midrule
\multirow{3}{*}{SDT}
 & ViT-S & \textbf{5.51}  \\
 & ViT-B & \textbf{9.45}  \\
 & ViT-L & \textbf{13.38} \\
\bottomrule
\end{tabular}
\end{wraptable}

we emphasize that our entire AnyDepth is designed from a light-weight and simple perspective, focusing not only on model design but also on data quality and quantity. 
Inspired by the principles of data-centric learning, we conclude that our model can achieve superior performance even with a relatively small amount of high-quality data (369K).

SDT uses only 5--13M parameters and outperforms DPT with various encoder sizes. 
Our results show that SDT significantly reduces the number of parameters and training cost while maintaining comparable accuracy to DPT, and there is a slight improvement in inference speed (Fig. \ref{fig:inference_time}). AnyDepth provides a lightweight, efficient, and computationally friendly alternative.

\subsection{Efficiency}
We comprehensively evaluated efficiency advantages of AnyDepth. Compared to DPT, AnyDepth not only significantly reduces the number of parameters (Fig.\ref{fig:model}), but also shows that AnyDepth significantly reduces FLOPs by 37\% when using models of varying sizes, particularly at high resolutions (Fig.\ref{fig:flops}). It also slightly improves inference speed (Fig.\ref{fig:inference_time}). Furthermore, Average iteration time of AnyDepth during training is 10\% shorter than that of DPT.

\begin{table}[t]
\centering
\caption{Inference latency comparison of SDT and DPT decoders on a Jetson Orin Nano (4GB).}
\label{tab:orin_nano_latency}
\begin{tabular}{l l r r}
\toprule
\textbf{Resolution} &
\textbf{Decoder} &
\textbf{Latency (ms)$\downarrow$} &
\textbf{FPS$\uparrow$} \\
\midrule
\multirow{2}{*}{256$\times$256}
 & DPT & 305.65 & 3.3 \\
 & \cellcolor[rgb]{.886,.937,.851}\textbf{SDT (Ours)}
 & \cellcolor[rgb]{.886,.937,.851}\textbf{213.35}
 & \cellcolor[rgb]{.886,.937,.851}\textbf{4.7} \\
\midrule
\multirow{2}{*}{512$\times$512}
 & DPT & 1107.64 & 0.9 \\
 & \cellcolor[rgb]{.886,.937,.851}\textbf{SDT (Ours)}
 & \cellcolor[rgb]{.886,.937,.851}\textbf{831.48}
 & \cellcolor[rgb]{.886,.937,.851}\textbf{1.2} \\
\bottomrule
\end{tabular}
\end{table}

To explore the sources of these efficiency improvements, we further compared the efficiency of the proposed SDT decoder and DPT decoder under the same experimental settings. As shown in Tables ~\ref{tab:decoder_params_vit} and Table ~\ref{tab:efficiency_multi_res}, SDT consistently and significantly reduces the number of parameters and computational cost across different ViT backbone network sizes and input resolutions. Importantly, the reduction in model size did not affect runtime performance, as the inference latency of SDT is comparable to or even slightly faster than that of DPT.

\begin{wrapfigure}{r}{0.5\textwidth}
\vspace{-1.3cm}
\centering
\includegraphics[width=0.5\textwidth]{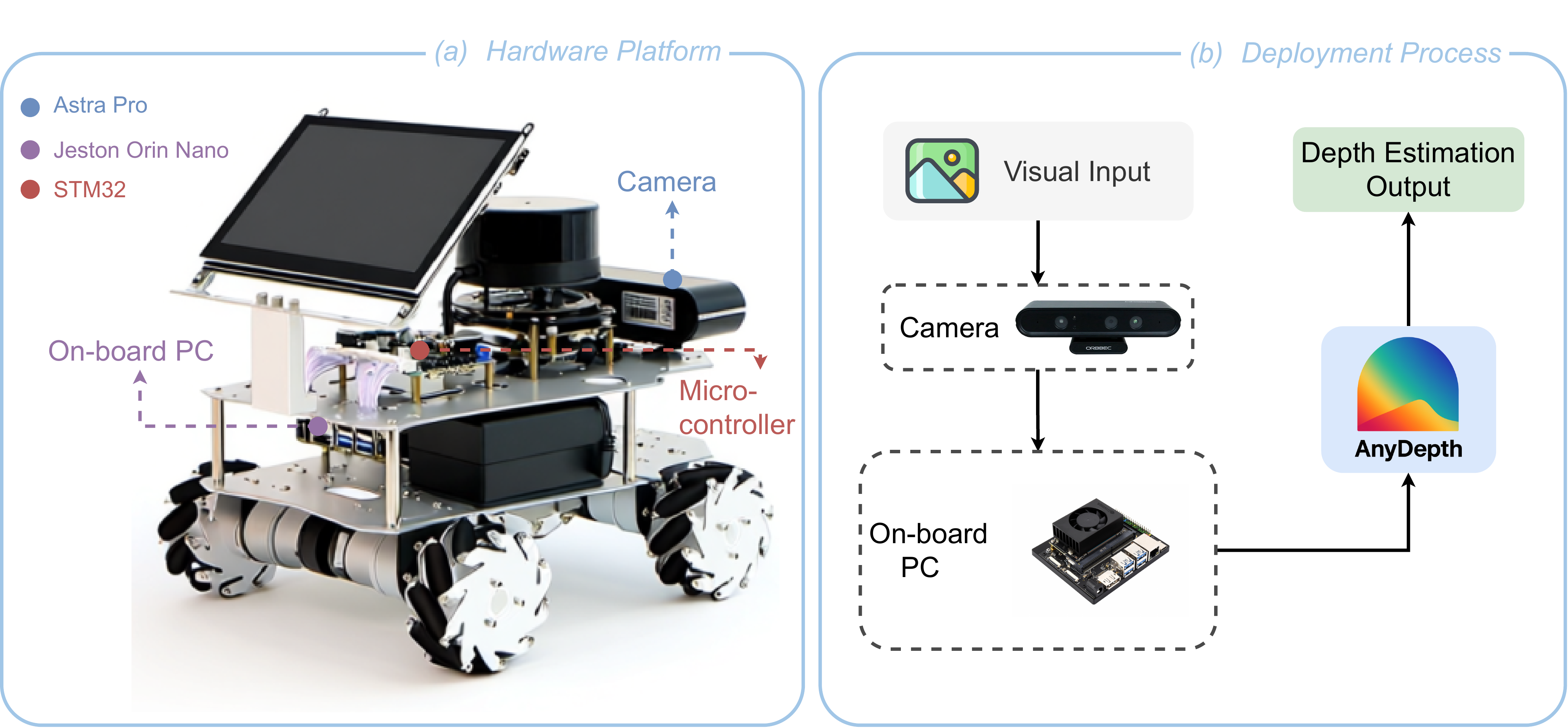}
\caption{Hardware and Evaluation Pipeline for Real-World Experiments}
\label{fig:real_world_overview}
    \vspace{-0.4cm}
\end{wrapfigure}

\subsection{Real World Evaluation}

As shown in Fig.~\ref{fig:real_world_overview}, We use the WHEELTEC R550 as the mobile platform for real-world evaluation. The robot is equipped with a Jetson Orin Nano 4GB as the onboard computing unit and an Astra Pro RGB-D camera as the perception unit. To evaluate its universality under various real-world conditions, we set up three different scenarios: a conference room, a corridor, and a rest area. Under the same encoder experimental setup, we used different decoders for real-world qualitative evaluation. As shown in Figure ~\ref{fig:real_world}, the SDT decoder performs better than the DPT decoder, displaying clearer boundaries in complex areas.
\begin{wraptable}{r}{0.45\textwidth}
\vspace{-0.5cm}
\centering
\caption{Peak GPU memory usage during inference at $256\times256$ resolution on Jetson Orin Nano (4GB).}
\label{tab:orin_nano_memory}
\small
\setlength{\tabcolsep}{6pt}
\renewcommand{\arraystretch}{1.0}
\begin{tabular}{l r}
\toprule
\textbf{Decoder} &
\textbf{Peak Memory (MB)$\downarrow$} \\
\midrule
DPT & 589.5 \\
\cellcolor[rgb]{.886,.937,.851}\textbf{SDT (Ours)}
 & \cellcolor[rgb]{.886,.937,.851}\textbf{395.2} \\
\bottomrule
\end{tabular}
\vspace{-0.5cm}
\end{wraptable}

Furthermore, we compared the efficiency performance of SDT and DPT on edge devices. As shown in Table ~\ref{tab:orin_nano_latency}, we compared the inference latency and throughput of the SDT and DPT decoders on the Jetson Orin Nano (4GB) at two input resolutions. At both 256×256 and 512×512 resolutions, SDT consistently outperforms DPT in terms of inference latency and frame rate. As shown in Table ~\ref{tab:orin_nano_memory}, at 256×256 resolution, SDT requires approximately 33\% less peak memory than the DPT decoder.

\subsection{Ablation Study}

We conducted ablation studies to validate our design. We used AnyDepth of ViT-B to progressively test our components, including data filtering, SDE, and DySample. As shown in the table~\ref{tab:ablation_filtering}, these ablation studies further support the effectiveness of data-centric learning in monocular depth estimation and demonstrate the detail enrichment capability of the SDE module and the additional gain of DySample compared to bilinear upsampling.

\section{Limitations and Future Work}
While our work demonstrates advantages, it also has some limitations. First, the current pipeline has not been evaluated in large-scale fully supervised or fine-tuned settings. Second, further analysis of the dataset can be used to optimize the filtering strategy. In future work, we can extend our lightweight framework to a wider range of tasks, such as metric depth and normal estimation.
\begin{table}[t]
\centering
\caption{Ablation experiments of AnyDepth-B on five benchmarks.
We report AbsRel (lower is better) and $\delta_1$ (higher is better).}
\label{tab:ablation_filtering}
\small
\setlength{\tabcolsep}{2pt}
\renewcommand{\arraystretch}{0.9}
\begin{tabular}{l *{5}{cc}}
\toprule
\multirow{2}{*}{\textbf{Method}} &
\multicolumn{2}{c}{\textbf{NYUv2}} &
\multicolumn{2}{c}{\textbf{KITTI}} &
\multicolumn{2}{c}{\textbf{ETH3D}} &
\multicolumn{2}{c}{\textbf{ScanNet}} &
\multicolumn{2}{c}{\textbf{DIODE}} \\
\cmidrule(lr){2-3}\cmidrule(lr){4-5}\cmidrule(lr){6-7}\cmidrule(lr){8-9}\cmidrule(lr){10-11}
& AbsRel$\downarrow$ & $\delta_1\uparrow$
& AbsRel$\downarrow$ & $\delta_1\uparrow$
& AbsRel$\downarrow$ & $\delta_1\uparrow$
& AbsRel$\downarrow$ & $\delta_1\uparrow$
& AbsRel$\downarrow$ & $\delta_1\uparrow$ \\
\midrule
w/o Filtering        & 9.5 & 91.1 & 15.4 & 77.3 & 14.0 & 91.2 & 8.3 & 93.5 & 25.0 & 71.1 \\
Filtering & 9.3 & 91.6 & 15.1 & 78.1 & 12.8 & 90.5 & 8.0 & 93.9 & 24.8 & 71.1 \\

Filtering + SDE & 8.8 & 92.4 & 14.7 & 79.6 & 11.5 & 91.0 & 7.9 & 94.1 & 24.3 & 71.1 \\

\cellcolor[rgb]{.886,.937,.851}\textbf{Filtering + SDE + Dysample}
& \cellcolor[rgb]{.886,.937,.851}\textbf{7.2}
& \cellcolor[rgb]{.886,.937,.851}\textbf{95.0}
& \cellcolor[rgb]{.886,.937,.851}\textbf{9.7}
& \cellcolor[rgb]{.886,.937,.851}\textbf{90.1}
& \cellcolor[rgb]{.886,.937,.851}\textbf{8.0}
& \cellcolor[rgb]{.886,.937,.851}\textbf{94.5}
& \cellcolor[rgb]{.886,.937,.851}\textbf{6.8}
& \cellcolor[rgb]{.886,.937,.851}\textbf{95.6}
& \cellcolor[rgb]{.886,.937,.851}\textbf{23.6}
& \cellcolor[rgb]{.886,.937,.851}\textbf{72.7} \\

\bottomrule
\end{tabular}
\end{table}

\section{Conclusion}
In this paper, we introduce AnyDepth, a simple and efficient-to-train framework for zero-shot monocular depth estimation. In our setup, a powerful self-supervised visual backbone paired with a single-path lightweight decoder is sufficient to achieve competitive performance without the need for large-scale, costly training. The goal of AnyDepth is not to surpass large-scale state-of-the-art methods, but rather to provide a more practical and academically valuable approach through its lightweight design and improved data quality.

\bibliography{iclr2026_conference}
\bibliographystyle{iclr2026_conference}

\clearpage
\appendix
\section{Appendix}

\subsection{LLM Use Declaration}
Large Language Models (ChatGPT) were used exclusively to improve the clarity and fluency of English writing. They were not involved in research ideation, experimental design, data analysis, or interpretation. The authors take full responsibility for all content.

\subsection{Data centric learning}
\label{sec:Data centric learning}

Although MiDaS~\citep{MiDaS} uses an affine-invariant loss to accommodate multi-dataset training, the varying degrees of noise and scale ambiguity introduced by these datasets can easily negatively impact training, especially in dense prediction tasks (Fig.\ref{fig:bad_samples}, \ref{fig:gradients}). Inspired by data-centric learning \citep{systematic, Data-centric}, for the monocular depth estimation task and our setting, we believe that high-quality samples should possess two properties: (i) depth values should be evenly distributed throughout the image, rather than being overly concentrated within a specific range; and (ii) gradient magnitudes should vary slightly across continuous surfaces, while exhibiting more pronounced changes near object edges. Based on these two properties, we define two metrics to measure sample quality. These metrics aim to reduce low-quality samples, facilitate model training, and reduce dataset size and training cost.

\subsubsection{Depth Distribution Score}
Some samples have depths that are primarily concentrated near or far, while other depth ranges are relatively small. As shown in Fig. \ref{fig:bad_samples} , this phenomenon is common in outdoor datasets. This unbalanced depth distribution can cause the model to favor learning depth values within a specific range rather than the entire valid depth range, leading to unstable training and poor model generalization.

To quantify this phenomenon, we propose a \textit{Depth Distribution Score} that evaluates      
how uniformly depth values are distributed across the available depth range. For a depth map ${D}\in\mathbb{R}^{H\times W}$, we divide the depth values into $K$ bins of equal width, and we use $K=20$ by default to balance granularity and robustness.

\textbf{Chi-square Deviation} ($S_{\chi^2}$). We measure the deviation from a uniform
distribution using the chi-square statistic:
\begin{equation}
\chi^2 = \sum_{k=1}^{K} \frac{(n_k - \bar{n})^2}{\bar{n}}, \quad S_{\chi^2} =
\exp\left(-\frac{\chi^2}{N}\right),
\label{eq:Chi-square Deviation}
\end{equation}
where $n_k$ is the number of depth bins $k$, $\bar{n} = N/K$ is the expected number under a uniform distribution, and $N$ is the total number of valid depth values. We use an exponential transformation to map the chi-squared statistic (Eq.~\ref{eq:Chi-square Deviation}) to $[0,1]$, with higher scores indicating a more uniform distribution.

\textbf{Maximum Concentration Index} ($S_{\text{conc}}$). To prevent excessive concentration in any single depth interval, we penalize the maximum bin occupancy:
\begin{equation}
S_{\text{conc}} = \begin{cases}
1, & \text{if } p_{\max} \leq 2/K \\
1 - \min\left(1, \frac{p_{\max} - 2/K}{0.5 - 2/K}\right), & \text{otherwise}
\end{cases}
\label{eq:Maximum Concentration Index}
\end{equation}
where $p_{\max} = \max_k(n_k)/N$ is the maximum bin probability. This formulation (Eq. \ref{eq:Maximum Concentration Index}) tolerates up to  twice the ideal concentration ($2/K$) without penalty, then linearly decreases the score as
concentration increases.

\paragraph{Range Utilization ($S_{\mathrm{range}}$)}.
Partition the available depth range into $K$ equal-width bins and let $n_k$ be the count in bin $k$.
Define the number of non-empty bins $K_+= \{\,k\in\{1,\dots,K\}\mid n_k>0\,\}$. The range utilization score is $S_{\mathrm{range}}={K_+}/{K}$, which penalizes samples whose depths concentrate within a narrow portion of the range.

The final Depth Distribution Score $S_{\text{dist}}$ is the weighted sum of these three scores:
\begin{equation}
S_{\text{dist}} = \lambda_1 \cdot S_{\chi^2} + \lambda_2 \cdot S_{\text{conc}} + \lambda_3 \cdot S_{\text{range}},
\label{eq:S_dist}
\end{equation}
where we empirically set $\lambda_1=0.5$, $\lambda_2=0.3$, and $\lambda_3=0.2$.

\subsubsection{Gradient Continuity Score}

In the real world, continuous physical surfaces should have smoothly transitioning depth values, without drastic random fluctuations. However, perhaps due to rendering defects in synthetic data, some sample depth maps exhibit gradient abrupt changes caused by noise on smooth surfaces. If these samples are used for training, the model will learn incorrect depth changes, thus affecting prediction quality.

Inspired by the gradient loss function (\citep{ECFNet, yang2018unsupervised, MiDaS}), we propose a \textit{gradient continuity score} to assess the noise content of each sample. We first calculate the gradient magnitude $G(i,j) = \sqrt{(\partial_x D)^2 + (\partial_y D)^2}$. To distinguish reasonable gradient abrupt changes at normal object edges from those caused by abnormal noise, we define edge pixels as pixels with gradient magnitudes in the top $10\%$. Within the smooth region, we use the coefficient of variation $\text{CV} = \frac{\sigma_G}{\mu_G}$ to assess gradient consistency:
\begin{equation}
S_{\text{grad}} = \frac{1}{1+\text{CV}},
\label{eq:S_grad}
\end{equation}
where $\mu_G$ and $\sigma_G$ are the mean and standard deviation of the gradient magnitude in the region, respectively.

\subsubsection{total score}
The depth distribution score and gradient continuity score capture different aspects of sample quality. We combine them into a \textit{Total Score}, defined as 
$S_{\text{total}} = (S_{\text{grad}} + S_{\text{dist}})/2$, 
to assess the overall quality of each sample for dataset filtering (Eq.~\ref{eq:S_dist}, \ref{eq:S_grad}). It's important to note that our goal is not to provide a particularly precise quality assessment method, but rather to design efficient indicators to quickly filter out samples with quality issues. For example, when performing edge detection, we did not use traditional Canny or Sobel algorithms because the detected edge maps often produce unnecessary artifacts and details. Learning-based methods, on the other hand, predict edges that are always several pixels off from their exact locations~\citep{ECFNet,he2019bi,pu2022edter,su2021pixel}, and their inference time is time-consuming, making them unsuitable for rapid filtering of large datasets.

\subsection{Visualization of low-quality samples}
\label{Visualization of low-quality samples}
\begin{figure}[htbp]
  \centering
  \includegraphics[width=\textwidth]{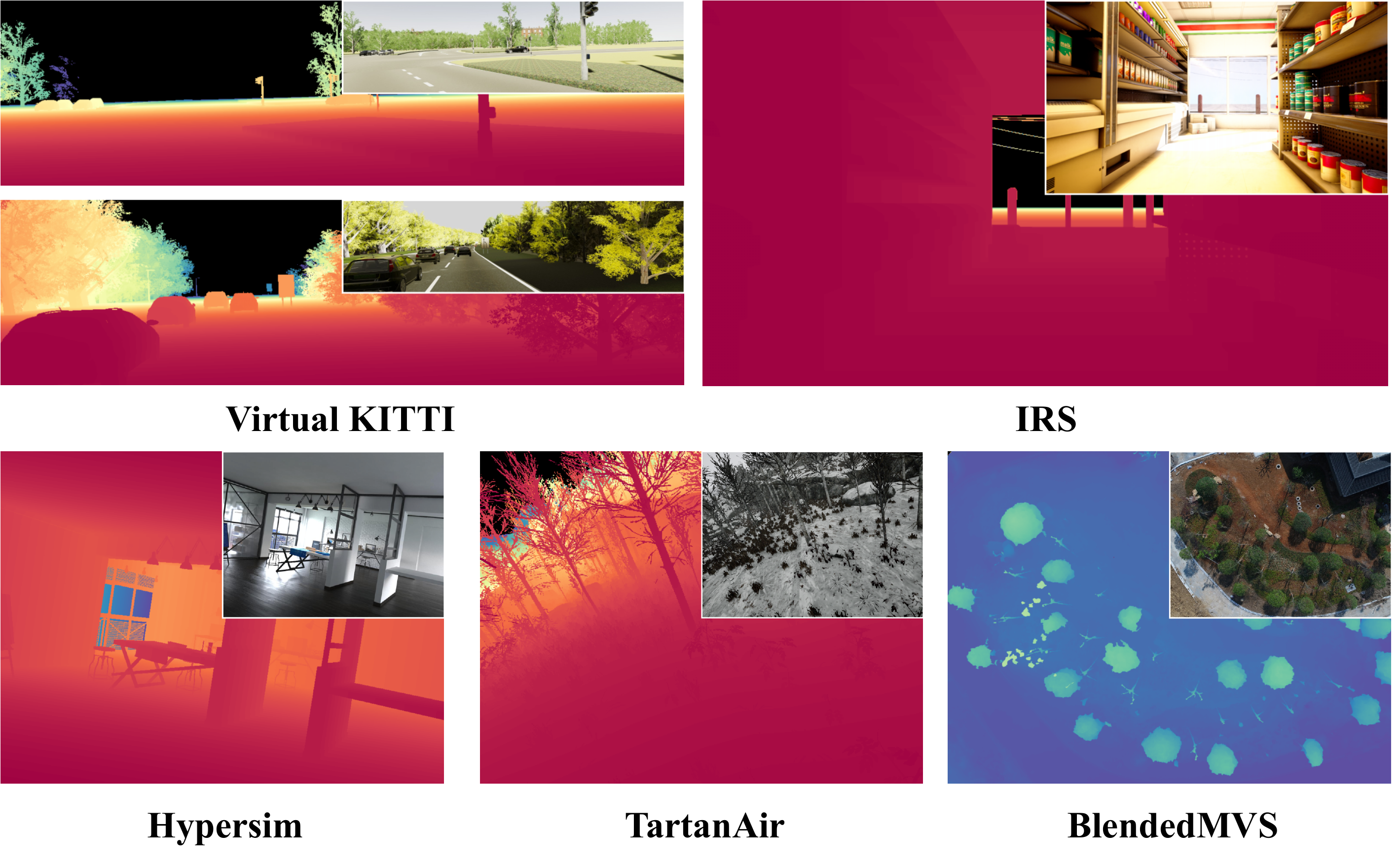}
  \caption{RGB images and GT of each dataset, showing that the depth value distribution of some samples is not uniform.}
  \label{fig:bad_samples}
\end{figure}

Figure~\ref{fig:bad_samples} provides qualitative examples of low-quality samples from five training datasets. It can be seen that some datasets contain samples with highly uneven depth value distributions, leading to biased supervision. This situation motivates us to use a depth distribution score when evaluating dataset quality.

In addition, Figure~\ref{fig:gradients} shows RGB images, gradient maps, and ground-truth depth examples from the same five datasets. The highlighted areas indicate the presence of severe gradient noise or inconsistent edges, which can negatively impact training stability. These qualitative findings support our quantitative gradient consistency metric.

\begin{figure}[t]
  \centering
  \includegraphics[width=\textwidth]{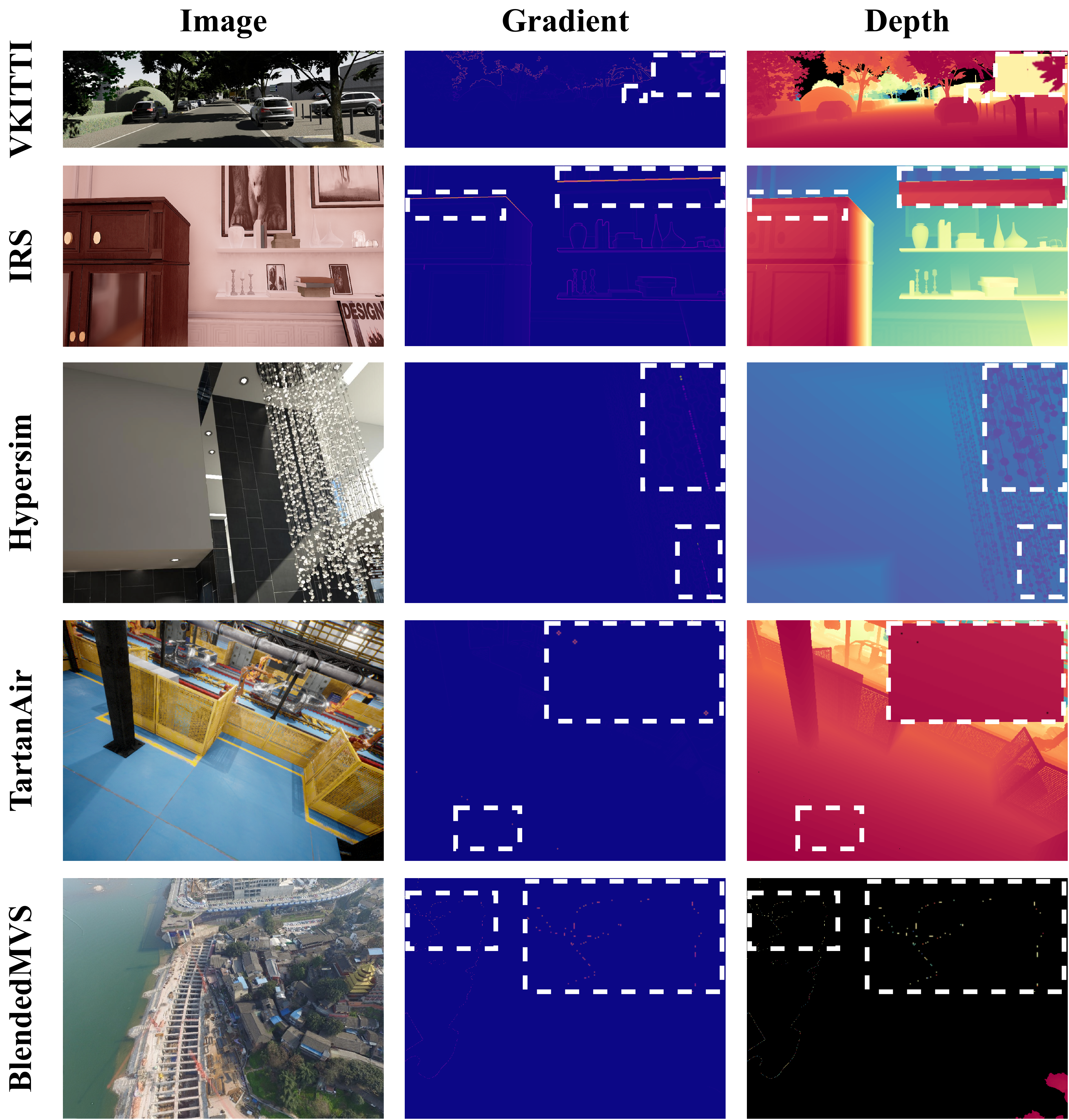}
  \caption{Examples of RGB, gradient, and GT depth from five datasets. 
  The dotted box highlights the noisy area.}
  \label{fig:gradients}
\end{figure}

\begin{figure}[t]
  \centering
  \includegraphics[width=\textwidth]{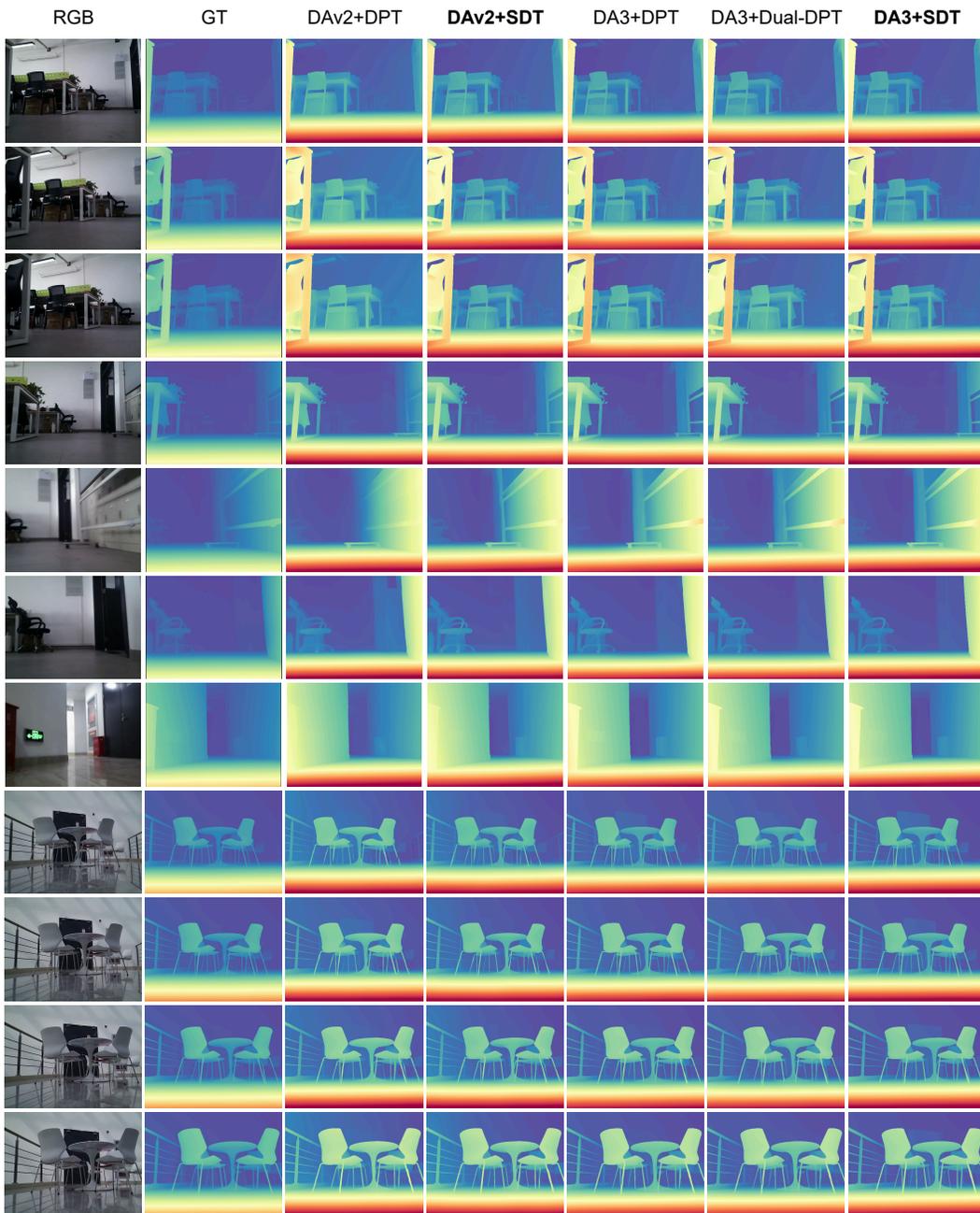}
  \caption{Qualitative results of zero-shot monocular depth estimation with different decoders (DPT, Dual-DPT, and SDT) using the same encoder.}
  \label{fig:real_world}
\end{figure}

\end{document}